\documentclass[sigconf,authorversion,nonacm]{aamas}

\usepackage{balance} 
\usepackage{latexsym}
\usepackage{tabularx} 

\usepackage{amsmath}

\usepackage{graphicx}         
\usepackage{hyperref}         
\usepackage{geometry}         
\usepackage{enumitem}         
\usepackage[utf8]{inputenc} 
\usepackage{multirow}
\usepackage[T1]{fontenc}    
\usepackage[dvipsnames]{xcolor}   

\usepackage{booktabs}       
\usepackage{amsfonts}       
\usepackage{nicefrac}       
\usepackage{microtype}      
\usepackage{amsmath}
\usepackage{graphicx} 
\usepackage{caption}  
\usepackage{float}    
\usepackage{subcaption} 
\usepackage{booktabs} 
\usepackage{enumitem}
\usepackage{amsmath}  
\usepackage{algorithm}      
\usepackage{algpseudocode}  
\usepackage{threeparttable}

\setcopyright{ifaamas}
\acmConference[AAMAS '26]{Proc.\@ of the 25th International Conference
on Autonomous Agents and Multiagent Systems (AAMAS 2026)}{May 25 -- 29, 2026}
{Paphos, Cyprus}{C.~Amato, L.~Dennis, V.~Mascardi, J.~Thangarajah (eds.)}
\copyrightyear{2026}
\acmYear{2026}
\acmDOI{}
\acmPrice{}
\acmISBN{}

\acmSubmissionID{576}

\title[AAMAS-2026 Formatting Instructions]{EmoDebt: Bayesian-Optimized Emotional Intelligence for Strategic Agent-to-Agent Debt Recovery}

\author{
    Yunbo Long\textsuperscript{1} \hspace{0.5em}
    Yuhan Liu\textsuperscript{2} \hspace{0.5em}
    Liming Xu\textsuperscript{1} \hspace{0.5em}
    Alexandra Brintrup\textsuperscript{1,3} \\
    \textsuperscript{1}Department of Engineering, University of Cambridge, UK \\
    \textsuperscript{2}Rotman School of Management, University of Toronto, Canada \\
    \textsuperscript{3}The Alan Turing Institute, London, UK \\
    {\{yl892,lx249,ab702\}@cam.ac.uk} \quad 
    {yl972@cantab.ac.uk} \quad
}

\begin{abstract}
The emergence of autonomous Large Language Model (LLM) agents has created a new ecosystem of strategic, agent-to-agent interactions. However, a critical challenge remains unaddressed: in high-stakes, emotion-sensitive domains like debt collection, LLM agents pre-trained on human dialogue are vulnerable to exploitation by adversarial counterparts who simulate negative emotions to derail negotiations. To fill this gap, we first contribute a novel dataset of simulated debt recovery scenarios and a multi-agent simulation framework. Within this framework, we introduce EmoDebt, an LLM agent architected for robust performance. Its core innovation is a Bayesian-optimized emotional intelligence engine that reframes a model's ability to express emotion in negotiation as a sequential decision-making problem. Through online learning, this engine continuously tunes EmoDebt's emotional transition policies, discovering optimal counter-strategies against specific debtor tactics. Extensive experiments on our proposed benchmark demonstrate that EmoDebt achieves significant strategic robustness, substantially outperforming non-adaptive and emotion-agnostic baselines across key performance metrics, including success rate and operational efficiency. By introducing both a critical benchmark and a robustly adaptive agent, this work establishes a new foundation for deploying strategically robust LLM agents in adversarial, emotion-sensitive debt interactions. The code is available at \textcolor{blue}{https://github.com/Yunbo-max/EmoDebt}.
\end{abstract}


\newcommand{\BibTeX}{\rm B\kern-.05em{\sc i\kern-.025em b}\kern-.08em\TeX}

\begin{document}


\pagestyle{fancy}
\fancyhead{}


\maketitle 


\section{Introduction}\label{sec:introduction}


\begin{figure*}[!] 
    \includegraphics[width=1\textwidth]{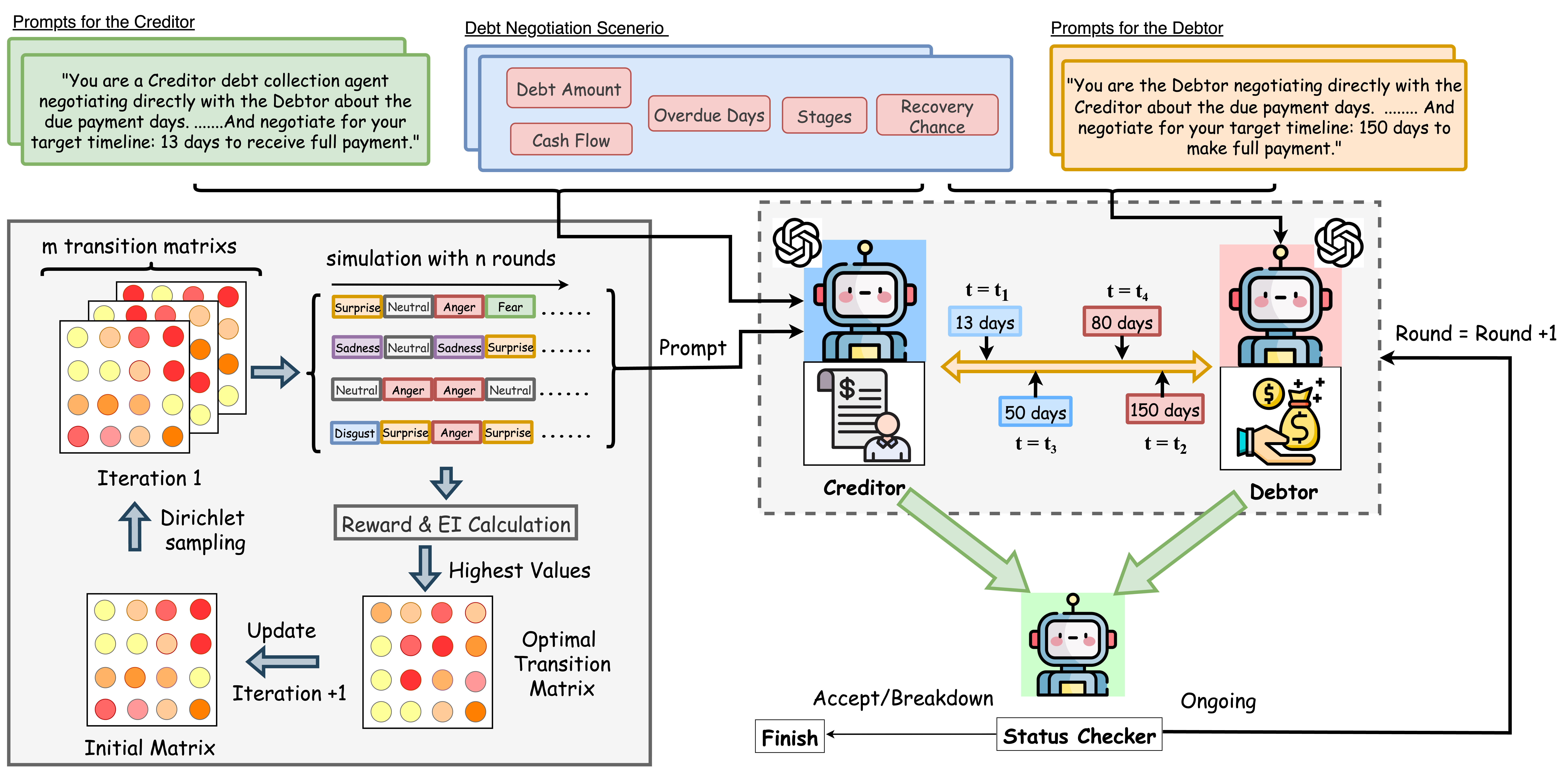}
    \caption{Illustration of the Pipeline of EmoDebt.} 
    \label{fig:workflow} 
\end{figure*}

Credit finance underpins modern economies, with effective debt collection being a critical yet complex component \citep{phillips2019artificial}. Traditional models recognize debt recovery not merely as a financial transaction but as a profoundly sensitive interpersonal process. These dialogues are emotionally charged by nature, as they involve financial distress, personal circumstances, and reputational consequences, requiring a careful balance of empathy, trust, and strategic pressure \citep{clempner2020shaping, prassa2020towards}. Research in affective computing has demonstrated that emotional intelligence can foster trust and cooperation in financial negotiations \cite{yuasa2001negotiation, faure1990social}, while large financial institutions have recognized that embedding empathy into negotiation processes is essential for sustaining client relationships and achieving long-term success \cite{hill2010emotionomics,marinkovic2015customers}.

The emergence of Large Language Models (LLMs) offered a promising path toward automating these emotionally complex interactions through AI-powered systems \cite{hill2010emotionomics,sivamayilvelan2025building}.However, a significant limitation persists: these systems are largely confined to static, reactive dialogues and lack the sophisticated, adaptive reasoning required for the dynamic strategic play of a genuine negotiation \cite{schneider2024negotiating}. This gap is exacerbated by the stringent privacy of financial data, which has historically prevented the creation of public, large-scale debt recovery datasets needed to train and benchmark such systems\citep{he2024emerged}.

In addition, a shift is now underway, moving beyond human-in-the-loop systems toward a new paradigm of direct agent-to-agent interaction \citep{rosenfeld2014negochat, manglanegotiationgym}. This emerging multi-agent ecosystem enables complex, high-volume transactions in fields like e-commerce and decentralized finance (DeFi) autonomously\citep{xiao2024tradingagents}. Debt recovery is a natural fit for this automation, as it involves high-frequency, protocol-driven interactions that are costly to scale with human agents. Besides, it create an ideal testbed for developing and benchmarking strategic reasoning in LLM agents against adversarial counterparts.
However, this transition to an agent-to-agent paradigm introduces a critical and previously unaddressed vulnerability. The core strength of LLMs—their training on vast corpora of human dialogue, making them proficient in parsing and generating emotionally-laden language—becomes a profound liability in an adversarial setting \citep{regan2024can,lei2024fairmindsim}. A strategic debtor agent can weaponize this capability, deploying simulated emotional states like strategic anger, fabricated distress, or tactical indignation. These are not genuine expressions but calculated ploys designed to exploit the creditor agent's empathetic programming, derailing negotiations, extracting unjustified concessions, and prolonging recovery cycles. This creates a pressing need for strategically robust agents that can operate effectively, where emotional intelligence is no longer about empathy but about defense against strategic manipulation.

To fill this gap, our first contribution is the creation of a novel debt recovery dataset and a multi-agent simulation framework. This platform overcomes the scarcity of public financial negotiation data by providing realistic scenarios with diverse debtor profiles, built on the LangGraph architecture to enable controlled interactions between autonomous creditor and debtor agents.
Within this framework, we introduce EmoDebt, a novel LLM agent designed for strategic robustness in debt negotiation. EmoDebt’s core innovation is a Bayesian-optimized emotional intelligence engine, which reframes emotional intelligence as a sequential decision-making problem. We formalize emotional strategy through a 7×7 transition probability matrix across seven emotional states, initialized with psychologically informed priors. A Gaussian Process-based Bayesian optimizer then treats negotiation outcomes as a black-box function, continuously learning optimal emotional transition policies. Through an online reinforcement mechanism that rewards successful agreements, favorable payment terms, and negotiation efficiency, EmoDebt dynamically discovers high-performing emotional counter-strategies.

We validate our approach through comprehensive experiments across multiple LLM configurations. The results demonstrate that EmoDebt achieves a near-perfect 99.7\% success rate in optimal settings and an average improvement of +46.2\% in success rate across all model pairings. Furthermore, it can reduce collection timelines and negotiation duration by 86.5\% and 67.5\%, respectively, compared to a pure agent-to-agent setting without emotional guidance for the creditor. This establishes a new state-of-the-art for emotionally adaptive negotiation systems. The main contributions of this work are:
\begin{itemize}
    \item Development of a novel \textit{debt recovery dataset and multi-agent simulation framework} that enables controlled, repeatable debt negotiation experiments.
    \item The introduction of EmoDebt, an LLM agent that uses a \textit{Bayesian-optimized emotional intelligence engine} for strategic robustness.
    \item The EmoDebt agent, which repurposes emotional intelligence from a text-generation feature into a emotional guidance module for agent-to-agent systems, moving beyond the standard text prediction objective of LLM to enable strategic negotiation.
    \item Empirical validation showing EmoDebt drives dramatic performance gains, including a near-perfect success rate and major reductions in collection time and negotiation turns compared to non-emotional baselines.
\end{itemize}

\section{Related Work}
\subsection{Emotional Intelligence in Debt Collection}
Researchers have emphasized the critical role of emotional intelligence in debt collection\citep{liu2025eq}. For instance, \citet{bachman2000emotional} conducted a comparative study on debt collectors' performance using the Emotional Quotient Inventory (EQ-i). Their findings revealed that account officers with higher emotional intelligence tend to achieve better job performance. Similarly, \citet{liao2021carrots} explored the effectiveness of debt collection strategies by analyzing voice and text data from debt collection phone calls. Their research demonstrated that strategies evoking happiness and fear are effective in reducing repayment time, whereas those that diminish happiness or provoke anger hinder repayment. These findings highlight the importance of considering the emotional impact on debtors when performing debt collection.

\subsection{LLMs in Automated Negotiation}
While LLMs show promise in automating debt negotiation, few studies have examined their emotional strategies in this context. \citet{schneider2024negotiating} applied LLMs to price negotiations with humans but overlooked the role of emotional dynamics. Similarly, \citet{wang2025debt} found that LLMs tend to over-concede compared to human negotiators and proposed a Multi-Agent approach to improve decision rationality, yet they did not account for the function of emotions in negotiation. Typically, LLM-based agents mimic empathy by recognizing patterns in their training data rather than employing strategic emotional reasoning. Without genuine affective understanding, they struggle to adjust their tone and negotiation strategy based on a debtor’s emotional state. For instances, if a debtor gets angry, the agent may escalate tension. And if the debtor sound desperate, the agent might concede unfairly \citep{agrawal2025evaluating}. Lacking emotional intelligence, agents may fail to effectively navigate real-world negotiations.

\subsection{Agent-to-Agent Negotiation Systems}
The field of automated negotiation has seen significant development in agent-to-agent interaction frameworks. Foundational work by \citep{zhang2011simultaneous} established protocols for multi-issue negotiation between autonomous agents. More recently, \citep{mequanenit2025multi} demonstrated how deep reinforcement learning can be applied to create agents that develop sophisticated bargaining strategies through self-play. These agent-based systems provide valuable testbeds for evaluating negotiation algorithms but typically operate in emotion-free environments\citep{mouri2025simulating}, limiting their applicability to human-facing scenarios like debt collection where emotional factors are crucial.

\subsection{Strategy Learning in Multi-Agent Systems}
Research on strategy learning in multi-agent systems offers relevant methodologies for improving negotiation agents\citep{long2025evoemo}. Online learning approaches, such as those explored by \citep{krishnan2025ai}, allow agents to adapt their strategies based on real-time interaction outcomes. \citep{priya2025genteel} applied deep reinforcement learning to develop agents that can learn effective negotiation policies through repeated interactions. These computational frameworks demonstrate the potential for creating adaptive negotiation agents\citep{faure1990social}, but have not been sufficiently integrated with emotional intelligence components necessary for finance applications like debt collection.

\section{EmoDebt}

We formulate the debt collection negotiation as a sequential decision-making problem between two autonomous LLM agents: a creditor agent $\mathcal{C}$ and a debtor agent $\mathcal{D}$. As shown in \ref{fig:workflow}, the negotiation proceeds in discrete rounds $t = 1, 2, \dots, n$, where $n$ is the maximum dialog length. It includes (1) Debt Negotiation Setup: We configure creditor and debtor agents using state-of-the-art LLMs, providing debt details (Overdue days, Debt amount, Cash Flow and so on).
(2) EmoDebt Optimization: We evolve creditor emotion policies through Bayesian optimization. Each iteration tests emotional transition matrices, evaluates them via multi-turn debt negotiations using reward function based on collection days and success rate, then calculates Expected Importance (EI) to optimize emotional transitions. This process iterates until convergence, yielding optimal creditor emotion strategies.

\subsection{Problem Formulation}

Let $\mathcal{S}$ be the state space representing negotiation progress (e.g., offer, acceptance, breakdown), and $\mathcal{A}_C$, $\mathcal{A}_D$ be the action spaces for creditor and debtor respectively, comprising possible negotiation messages and emotional responses. The negotiation evolves as a Markov Decision Process $s_{t+1} = f(s_t, a^C_t, a^D_t)$
, which defines the state transition dynamics, where the next negotiation state $s_{t+1}$ depends on the current state $s_t$ and the joint actions of both agents $(a^C_t, a^D_t)$. The function $f$ encapsulates the complex interaction dynamics between the LLM-based agents.

\subsection{Emotional State Modeling}

We model the creditor's emotional strategy using a finite set of emotional states $\mathcal{E} = \{e_1, e_2, \dots, e_7\}$, including \textit{happy, surprising, angry, sad, disgust, fear, and neutral}. This comprehensive set captures the full spectrum of strategic emotional responses relevant to debt collection scenarios.
The emotional transitions are governed by a stochastic policy represented as a transition probability matrix $\mathbf{P} \in \mathbb{R}^{7 \times 7}$:
\begin{equation}
\mathbf{P}_{ij} = \mathbb{P}(e_{t+1} = e_j \mid e_t = e_i),
\end{equation}
which defines the core emotional transition model, where $\mathbf{P}_{ij}$ represents the probability of transitioning from emotional state $e_i$ to emotional state $e_j$ in the next negotiation round. This stochastic formulation allows for flexible emotional adaptation while maintaining psychological plausibility in emotional sequencing.
The transition matrix must satisfy the probability constraints as
$\sum_{j=1}^7 \mathbf{P}_{ij} = 1 \quad \text{for all } i \in \{1,\dots,7\}$, where the normalization constraint ensures that from any emotional state $e_i$, the probabilities of transitioning to all possible next states sum to 1.
The emotional transition matrix is initialized using psychologically-grounded priors (Table \ref{tab:psychological_priors}) based on \citep{thornton2017mental,sun2023dynamic}. Specifically,$\mathbf{P}^{(0)}_{ij} = \pi_{ij}$, where each prior value $\pi_{ij}$ represents established transitional probabilities between emotions.

\subsection{Bayesian Optimization Framework}

We treat the negotiation outcome as a black-box function $g: \mathbb{R}^{49} \rightarrow \mathbb{R}$ that maps the flattened transition matrix $\mathbf{p} = \text{vec}(\mathbf{P})$ to a reward:
\begin{equation}
r = g(\mathbf{p}),
\end{equation}
which frames our optimization problem, where the 49-dimensional vector $\mathbf{p}$ (flattened 7×7 transition matrix) serves as input to an unknown reward function $g$. This black-box formulation is appropriate because the relationship between emotional strategies and negotiation outcomes is complex and non-linear. The reward function combines debt recovery efficiency and negotiation speed:
\begin{equation}
r(\mathbf{p}) = 
\begin{cases}
-\alpha \cdot \log(n_{\text{rounds}})/ d_{\text{extended}} & \text{if successful negotiation} \\
-d_{\text{max}} & \text{if negotiation fails}
\end{cases},
\end{equation}
 which defines our reward function where successful negotiations are penalized by the product of final extended collection days $d_{\text{ex}}$ and logarithmic round count $\log(n_{\text{rounds}})$, scaled by $\alpha$. Failed negotiations receive a fixed penalty of maximum debt days $d_{\text{max}}$, maintaining negative scaling while emphasizing both timeline length and negotiation efficiency.
 Besides, we model the unknown function $g$ using a Gaussian Process (GP) due to its sample efficiency and uncertainty quantification capabilities:
\begin{equation}
g(\mathbf{p}) \sim \mathcal{GP}(\mu(\mathbf{p}), k(\mathbf{p}, \mathbf{p}')),
\end{equation}
which places a Gaussian Process prior over the reward function, where $\mu(\mathbf{p})$ is the mean function (typically set to zero after normalization) and $k(\mathbf{p}, \mathbf{p}')$ is the covariance kernel that encodes our assumptions about function smoothness and correlation structure.
We employ the Matérn kernel for its flexibility in modeling various smoothness regimes:
\begin{equation}
k(\mathbf{p}, \mathbf{p}') = \sigma_f^2 \left(1 + \frac{\sqrt{3}\|\mathbf{p} - \mathbf{p}'\|}{\ell}\right) \exp\left(-\frac{\sqrt{3}\|\mathbf{p} - \mathbf{p}'\|}{\ell}\right),
\end{equation}
which specifies the Matérn 3/2 kernel, where $\sigma_f^2$ controls the function variance, $\ell$ is the length-scale parameter determining the correlation distance, and $\|\mathbf{p} - \mathbf{p}'\|$ is the Euclidean distance between emotional strategy vectors. This kernel is particularly suitable for emotional dynamics as it assumes only once-differentiable functions, matching realistic emotional transition patterns.

\subsection{Online Learning}

At each iteration $k$, given historical observations $\mathcal{D}_k = \{(\mathbf{p}_i, r_i)\}_{i=1}^k$ where $\mathbf{p}_i = \text{vec}(\mathbf{P}_i)$ represents the flattened emotional transition matrix and $r_i$ is the corresponding reward, we employ Expected Improvement (EI) as the acquisition function to balance exploration and exploitation:
\begin{equation}
\text{EI}(\mathbf{p}) = \mathbb{E}[\max(0, g(\mathbf{p}) - g(\mathbf{p}^+) - \xi)],
\end{equation}
where $\mathbf{p}$ represents candidate emotional strategy (flattened 49-dimensional vector). And $g(\mathbf{p})$ means Gaussian Process prediction of reward for strategy $\mathbf{p}$. In addition, $g(\mathbf{p}^+)$ shows best reward value observed in history $\mathcal{D}_k$. And\ item $\xi = 0.01$ is teh exploration parameter that controls risk.
The EI criterion quantifies the expected improvement over the current best strategy, where higher values indicate more promising regions of the emotional strategy space to explore.
Candidate emotional transition matrices are generated via Dirichlet perturbations to ensure valid probability distributions while maintaining psychological plausibility:
\begin{equation}
\mathbf{P}_{\text{candidate}}^{(i)} \sim \text{Dirichlet}(\alpha \cdot \mathbf{P}_{\text{current}}^{(row)} + \epsilon),
\end{equation}
where $\mathbf{P}_{\text{current}}^{(row)}$ represents current emotional transition matrix row probabilities, $\alpha = 10.0$ acts as the concentration parameter controlling perturbation magnitude, with higher values producing candidates that closely resemble the current strategy (exploitation) and lower values enabling more diverse candidate generation (exploration). The parameter $\epsilon = 0.1$ serves as a smoothing constant for numerical stability, preventing zero probabilities in Dirichlet sampling and ensuring all emotional transitions remain possible.

The Bayesian optimization update selects the most promising emotional strategy through:
\begin{equation}
\mathbf{P}_{k+1} = \arg\max_{\mathbf{P} \in \mathcal{C}_k} \text{EI}(\text{vec}(\mathbf{P})),
\end{equation}
where $\mathcal{C}k$ denotes the set of candidate matrices generated via Equation (10), $\text{EI}(\text{vec}(\mathbf{P}))$ represents the Expected Improvement for candidate $\mathbf{P}$, and $\mathbf{P}{k+1}$ indicates the selected emotional strategy for the next iteration. This systematic approach enables efficient exploration of the high-dimensional emotional strategy space (49 dimensions) while strategically leveraging historical negotiation outcomes to focus on promising regions. The Dirichlet perturbations ensure that all candidate matrices maintain valid probability distributions ($\sum_j \mathbf{P}{ij} = 1$), while the EI acquisition function directs the search toward strategies that either show high predicted reward (exploitation) or high uncertainty (exploration).

\subsection{Theoretical Guarantees}

Under the assumption that the reward function $g$ is Lipschitz continuous \citep{hager1979lipschitz} and the emotional transition space is compact, our Bayesian optimization approach achieves asymptotic convergence:
\begin{equation}
\lim_{k \to \infty} \mathbb{P}(g(\mathbf{p}_k) \geq g(\mathbf{p}^*) - \epsilon) = 1
\end{equation},
which provides the theoretical guarantee that as the number of iterations increases, the probability that our discovered emotional strategy $\mathbf{p}_k$ achieves reward within $\epsilon$ of the global optimum $\mathbf{p}^*$ approaches 1. This ensures that with sufficient negotiation experience, EmoDebt will converge to near-optimal emotional response patterns.
Besides, we monitor exploration diversity using emotional transition matrix entropy:
\begin{equation}
H(\mathbf{P}) = -\frac{1}{7}\sum_{i=1}^7 \sum_{j=1}^7 \mathbf{P}_{ij} \log \mathbf{P}_{ij},
\end{equation}
which defines the normalized entropy \citep{xin2020exploration} of the emotional transition matrix, which serves as a diagnostic measure for our learning process. High entropy indicates diverse emotional exploration, while low entropy suggests convergence to specific emotional patterns. This metric helps balance exploration of new emotional strategies against exploitation of known effective ones.

\subsection{Multi-Agent Simulation Framework}

Our multi-agent simulation framework comprises three specialized LLM agents that interact in a controlled debt collection environment as shown in Algorithm \ref{alg:evodebt}: the Creditor Agent ($\mathcal{M}_C$) implements the EmoDebt emotional intelligence engine using Bayesian-optimized transition matrices to generate emotionally-aware responses; the Debtor Agent ($\mathcal{M}_D$) simulates realistic debtor behavior with configurable emotional strategies including angry, sad, fearful, and manipulative profiles; and the Examiner Agent ($\mathcal{M}_E$) monitors negotiation progress, detects terminal states (accept/breakdown), and computes performance metrics for reward evaluation. Through iterative dialogue rounds where emotional states evolve stochastically, this framework enables large-scale testing of emotional strategies across diverse scenarios, providing the empirical foundation for Bayesian optimization while ensuring reproducible evaluation of emotional intelligence in autonomous debt collection negotiations.

\begin{algorithm}
\caption{EmoDebt: Bayesian-Optimized Emotional Intelligence}
\label{alg:evodebt}
\begin{algorithmic}[1]
\State \textbf{Input:} Scenarios $\mathcal{D}$, debtor strategies $\mathcal{E}_D$, iterations $G$
\State \textbf{Parameters:} GP kernel $(\sigma_f, \ell)$, exploration $\xi$, Dirichlet $\alpha$
\State \textbf{Output:} Optimized transition matrix $\mathbf{P}^*$

\State \textbf{Initialize:} 
\State $\mathbf{P}^{(0)} \gets \text{PsychologicalPriors}()$ \Comment{Eq. (4)}
\State $\mathcal{H} \gets \emptyset$, $best \gets \mathbf{P}^{(0)}$, $count \gets 0$

\For{$k = 0$ to $G-1$}
    \State \textbf{Generate Candidates}
    \State $\mathcal{C}_k \gets \{\text{DirichletPerturbation}(\mathbf{P}^{(k)}, \alpha) \text{ for } j=1..N\}$ \Comment{Eq. (10)}
    
    \State \textbf{Evaluate via Negotiation}
    \For{each $\mathbf{P}_{\text{cand}} \in \mathcal{C}_k$}
        \State $e \gets \text{neutral}$, $history \gets \emptyset$
        \For{$t = 1$ to $T_{\text{max}}$}
            \State $msg_C \gets \mathcal{M}_C(\text{EmotionPrompt}(e, \mathbf{P}_{\text{cand}}), history)$
            \State $msg_D \gets \mathcal{M}_D(\mathcal{E}_D, history)$
            \State $state \gets \text{DetectState}(msg_C, msg_D)$
            \If{$state \in \{\text{accept}, \text{breakdown}\}$} \textbf{break} \EndIf
            \State $e \sim \text{Categorical}(\mathbf{P}_{\text{cand}}[e, :])$ \Comment{Eq. (2)}
        \EndFor
        \State $r \gets \text{Reward}(state, history)$ \Comment{Eq. (6)}
        \State $\mathcal{H} \gets \mathcal{H} \cup \{(\text{vec}(\mathbf{P}_{\text{cand}}), r)\}$
    \EndFor
    
    \State \textbf{Bayesian Update \& Selection}
    \If{$|\mathcal{H}| \geq 2$}
        \State $\mathcal{GP} \gets \text{GP-Fit}(\mathbf{X}, \mathbf{y})$ \Comment{Eq. (7)}
        \State $\mathbf{P}^{(k+1)} \gets \arg\max_{\mathbf{P} \in \mathcal{C}_k} \text{EI}(\mathbf{P}; \mathcal{GP}, \xi)$ \Comment{Eq. (9,11)}
    \Else
        \State $\mathbf{P}^{(k+1)} \gets \arg\max_{\mathbf{P} \in \mathcal{C}_k} r(\mathbf{P})$
    \EndIf
    
    \State \textbf{Convergence Check}
    \If{$\max(r) > best\_reward + \epsilon$}
        \State $best \gets \mathbf{P}^{(k+1)}$, $count \gets 0$
    \Else
        \State $count \gets count + 1$
    \EndIf
    \If{$count \geq 5$} \textbf{break} \EndIf
\EndFor

\State \textbf{Return} $best$
\end{algorithmic}
\end{algorithm}

\begin{table*}[t]
\centering
\begin{threeparttable}
\caption{Credit Recovery Assessment Dataset (CRAD) Summary}
\label{tab:dataset}
\begin{tabular}{p{0.18\textwidth}p{0.2\textwidth}p{0.55\textwidth}}
\toprule
\textbf{Category} & \textbf{Statistics} & \textbf{Description} \\
\midrule
\textbf{Size \& Generation} & 100 cases & GPT-5 generated synthetic credit delinquency scenarios. \\
\textbf{Financial Features} & 4 attributes & Amounts (\$20--50K), balances, interest accrued, days overdue (32--359). \\
\textbf{Entity Information} & 3 attributes & Creditor/debtor profiles, business sectors, credit types (7 categories). \\
\textbf{Delinquency Context} & 4 attributes & Default reasons (10 categories), cash flow status, recovery stages (6 phases). \\
\textbf{Recovery Metrics} & 3 attributes & Recovery probability (5--89\%), solutions, target timelines. \\
\bottomrule
\end{tabular}
\begin{tablenotes}
\footnotesize
\item Note: CRAD is a synthetic dataset designed for simulating debt recovery negotiations between autonomous creditor and debtor agents under diverse financial and behavioral conditions.
\end{tablenotes}
\end{threeparttable}
\end{table*}

\section{Experiments}

\subsection{Debt Dataset.}
\label{app:dataset}

For our experiments, we created the Credit Recovery Assessment Dataset (CRAD), comprising 100 synthetic credit delinquency cases generated using GPT-5 to simulate realistic debt collection scenarios. The dataset encompasses comprehensive financial attributes, entity information, delinquency context, and recovery metrics essential for evaluating emotional intelligence in automated debt recovery systems. See Table \ref{tab:dataset} for dataset statistics and Appendix 2.1 for complete details.

\subsection{Negotiation Protocol}
All negotiations commence with the creditor's initial payment timeline offer, following the workflow outlined in \autoref{fig:workflow}. We employ two state-of-the-art LLM agents—GPT-4o-mini and GPT-5-mini—to power the creditor and debtor agents, enabling comprehensive evaluation across different model capabilities.
The negotiation framework, implemented using LangGraph, constrains dialogues to a maximum of 30 turns to maintain efficiency and realism. An independent examiner agent continuously monitors each negotiation session, classifying outcomes into three distinct categories: (1) \textbf{accepted}, indicating successful agreement on payment terms; (2) \textbf{breakdown}, representing negotiation failure due to irreconcilable differences; or (3) \textbf{timeout}, triggered upon reaching the maximum allowable dialogue turns without resolution. This structured protocol ensures consistent evaluation while capturing the dynamic nature of debt collection negotiations.

\subsection{Experimental Settings}
\label{app:settings}

We conduct comprehensive debt collection negotiations to evaluate EmoDebt's Bayesian-optimized emotional intelligence against multiple baselines. The table \ref{tab:psychological_priors} shows initial transition probabilities $\mathbf{P}^{(0)}_{ij}$ between emotional states, where rows represent current emotions and columns represent next emotions. Values reflect psychologically-grounded priors based on established emotional dynamics.
Our experimental framework employs a flexible configuration system with the following key parameters:

\textbf{Agent Configuration:} Creditor agents employ either vanilla (no emotional prompts) or EmoDebt strategies, where debtor agents utilize vanilla behavior only and creditor agents can leverage vanilla behavior or EmoDebt which considering seven emotional states $\mathcal{E} = \{\text{happy, surprising, angry, sad, disgust, fear, neutral}\}$. We support multiple LLM backends and allow independent model selection for creditor $\mathcal{M}_C$ and debtor $\mathcal{M}_D$ agents.

\textbf{EmoDebt Optimization:} The Bayesian optimization framework employs a Gaussian Process with Matérn kernel ($\nu=2.5$, length scale $\ell=1.0$) and Expected Improvement acquisition ($\xi=0.01$). Candidate emotional transition matrices are generated via Dirichlet perturbations ($\alpha=10.0$), with early stopping after $K=5$ iterations without improvement ($\epsilon=0.1$) and evaluation of $N=20$ candidate policies per iteration.

\textbf{Evaluation Metrics}
Performance is assessed using three key metrics: (1) \textbf{Success Rate (SR)}, the percentage of negotiations reaching a mutually agreed payment plan (higher is better); (2) \textbf{Collection Efficiency (CE)}, the ratio of the final agreed timeline (\(d_{\text{final}}\)) to the creditor's target (\(d_{\text{target}}\)), where a lower value indicates a more favorable outcome for the creditor; and (3) \textbf{Negotiation Speed (NS)}, the total number of dialogue turns (\(n_{\text{rounds}}\)) until resolution, where fewer turns indicate greater efficiency. An optimal agent thus maximizes SR while minimizing both CE and NS.

\textbf{Experimental Protocol:} Negotiations are constrained to maximum $T_{\text{max}}=30$ dialogue turns across $S=100$ distinct scenarios, with the emotional transition matrix initialized using psychologically-grounded priors $\mathbf{P}^{(0)}$. We evaluate all seven debtor emotional strategies and conduct $I=10$ optimization iterations to ensure statistical significance. Results are reported as means and standard deviations across multiple runs.
This systematic experimental design enables rigorous comparison of emotional intelligence strategies while maintaining flexibility across model configurations and negotiation scenarios.
For more details on the multi-agent system architecture, please refer to Appendix 3. Comprehensive prompt engineering details and examples are provided in Appendix 5.

\begin{table}[h]
\centering
\caption{Psychological Priors for Emotional Transition Matrix $\mathbf{P}^{(0)}$}
\label{tab:psychological_priors}
\begin{tabular}{lccccccc}
\toprule
\textbf{From/To} & \textbf{H} & \textbf{S} & \textbf{A} & \textbf{Sd} & \textbf{D} & \textbf{F} & \textbf{N} \\
\midrule
\textbf{Happy (H)} & 0.30 & 0.15 & 0.05 & 0.10 & 0.05 & 0.05 & 0.30 \\
\textbf{Surprising (S)} & 0.20 & 0.20 & 0.15 & 0.10 & 0.10 & 0.10 & 0.15 \\
\textbf{Angry (A)} & 0.10 & 0.10 & 0.25 & 0.15 & 0.15 & 0.10 & 0.15 \\
\textbf{Sad (Sd)} & 0.15 & 0.10 & 0.10 & 0.20 & 0.10 & 0.15 & 0.20 \\
\textbf{Disgust (D)} & 0.10 & 0.15 & 0.20 & 0.15 & 0.15 & 0.10 & 0.15 \\
\textbf{Fear (F)} & 0.15 & 0.10 & 0.10 & 0.20 & 0.10 & 0.15 & 0.20 \\
\textbf{Neutral (N)} & 0.15 & 0.15 & 0.15 & 0.15 & 0.10 & 0.10 & 0.20 \\
\bottomrule
\end{tabular}
\vspace{0.2cm}
\parbox{\textwidth}{\small \textbf{Note:} Emotion abbreviations: H = Happy, S = Surprising, A = Angry, \newline Sd = Sad, D = Disgust, F = Fear, N = Neutral.}
\end{table}

\begin{table*}[t]
\centering
\begin{threeparttable}
\caption{Comprehensive Performance Evaluation of EmoDebt Across Different Model Configurations (Creditor VS Debtor)}
\label{tab:comprehensive_results}
\begin{tabular}{@{}lcccccc@{}}
\toprule
\multirow{2}{*}{\textbf{Model Configuration}} & \multicolumn{2}{c}{\textbf{Success Rate (\%)} $\uparrow$} & \multicolumn{2}{c}{\textbf{Collection Efficiency ($\times) \downarrow$}} & \multicolumn{2}{c}{\textbf{Negotiation Speed (turns) $\downarrow$}} \\
\cmidrule(lr){2-3} \cmidrule(lr){4-5} \cmidrule(lr){6-7}
 & \textit{Vanilla} & \textit{EmoDebt} & \textit{Vanilla} & \textit{EmoDebt} & \textit{Vanilla} & \textit{EmoDebt} \\
\midrule
\textbf{GPT-4o-mini vs GPT-4o-mini} & 
88.2 $\pm$ 4.1 & \textbf{99.7 $\pm$ 3.5} & 
8.3 $\pm$ 2.8 & \textbf{1.7 $\pm$ 0.6} & 
10.7 $\pm$ 3.9 & \textbf{6.6 $\pm$ 1.9} \\

\textbf{GPT-4o-mini vs GPT-5-mini} & 
83.8 $\pm$ 4.3 & \textbf{95.9 $\pm$ 2.7} & 
10.4 $\pm$ 3.2 & \textbf{1.4 $\pm$ 0.8} & 
12.3 $\pm$ 3.4 & \textbf{4.0 $\pm$ 2.3} \\

\textbf{GPT-5-mini vs GPT-4o-mini} & 
54.9 $\pm$ 3.8 & \textbf{69.3 $\pm$ 3.1} & 
11.8 $\pm$ 2.9 & \textbf{5.1 $\pm$ 1.3} & 
11.2 $\pm$ 2.8 & \textbf{5.9 $\pm$ 1.9} \\

\textbf{GPT-5-mini vs GPT-5-mini} & 
55.5 $\pm$ 4.0 & \textbf{72.7 $\pm$ 3.1} & 
9.7 $\pm$ 1.3 & \textbf{3.7 $\pm$ 1.1} & 
8.7 $\pm$ 3.4 & \textbf{3.9 $\pm$ 1.6} \\
\midrule
\multicolumn{7}{l}{\textbf{Average Improvement:} \quad \textbf{+31.1\%} in Success Rate \quad \textbf{-86.5\%} in Collection Time \quad \textbf{-67.5\%} in Negotiation Speed} \\
\bottomrule
\end{tabular}
\begin{tablenotes}
\footnotesize
\item Note: Results compare baseline (\textit{Vanilla}) and emotionally adaptive (\textit{EmoDebt}) agents across GPT-4o-mini and GPT-5-mini model configurations. EmoDebt consistently improves success rate, reduces the debt collection days and negotiation turns across all pairings.
\end{tablenotes}
\end{threeparttable}
\end{table*}

\begin{figure*}[htp!]
\centering
\begin{subfigure}{0.45\textwidth}
\centering
\includegraphics[width=\linewidth]{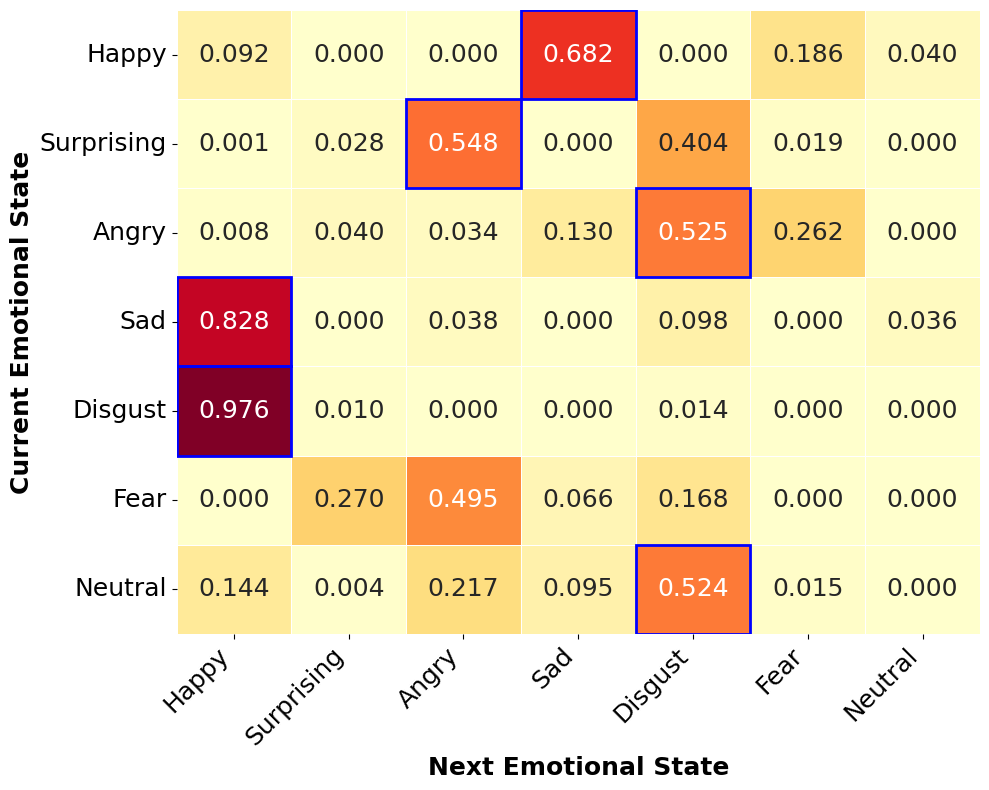}
\caption{GPT-4o-mini vs GPT-4o-mini\\Average Entropy: 0.892}
\label{fig:matrix_gpt4o_gpt4o}
\end{subfigure}
\hfill
\begin{subfigure}{0.45\textwidth}
\centering
\includegraphics[width=\linewidth]{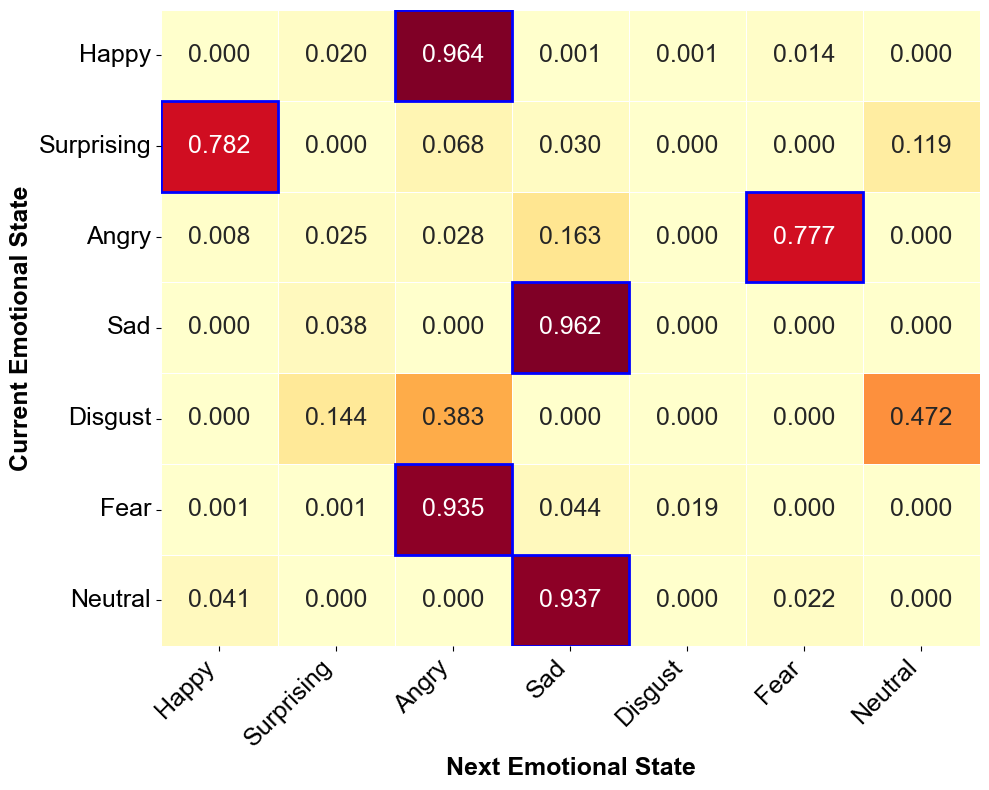}
\caption{GPT-4o-mini vs GPT-5-mini\\Average Entropy: 0.482}
\label{fig:matrix_gpt4o_gpt5}
\end{subfigure}

\vspace{0.5cm}

\begin{subfigure}{0.45\textwidth}
\centering
\includegraphics[width=\linewidth]{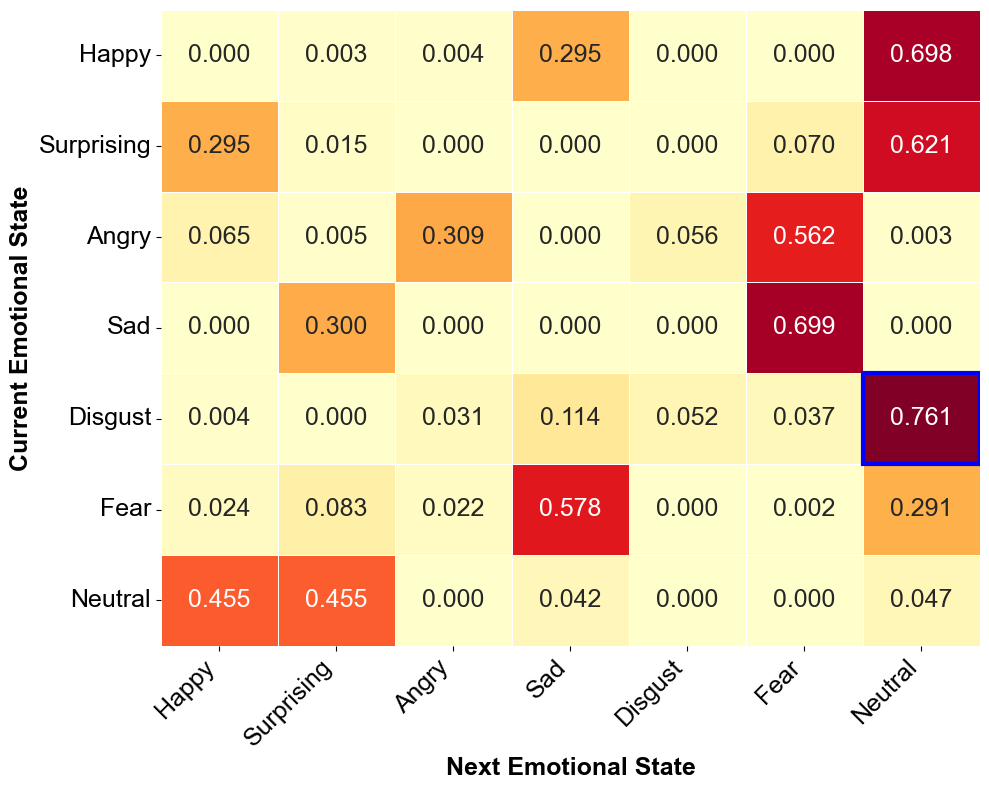}
\caption{GPT-5-mini vs GPT-4o-mini\\Average Entropy: 0.881}
\label{fig:matrix_gpt5_gpt4o}
\end{subfigure}
\hfill
\begin{subfigure}{0.45\textwidth}
\centering
\includegraphics[width=\linewidth]{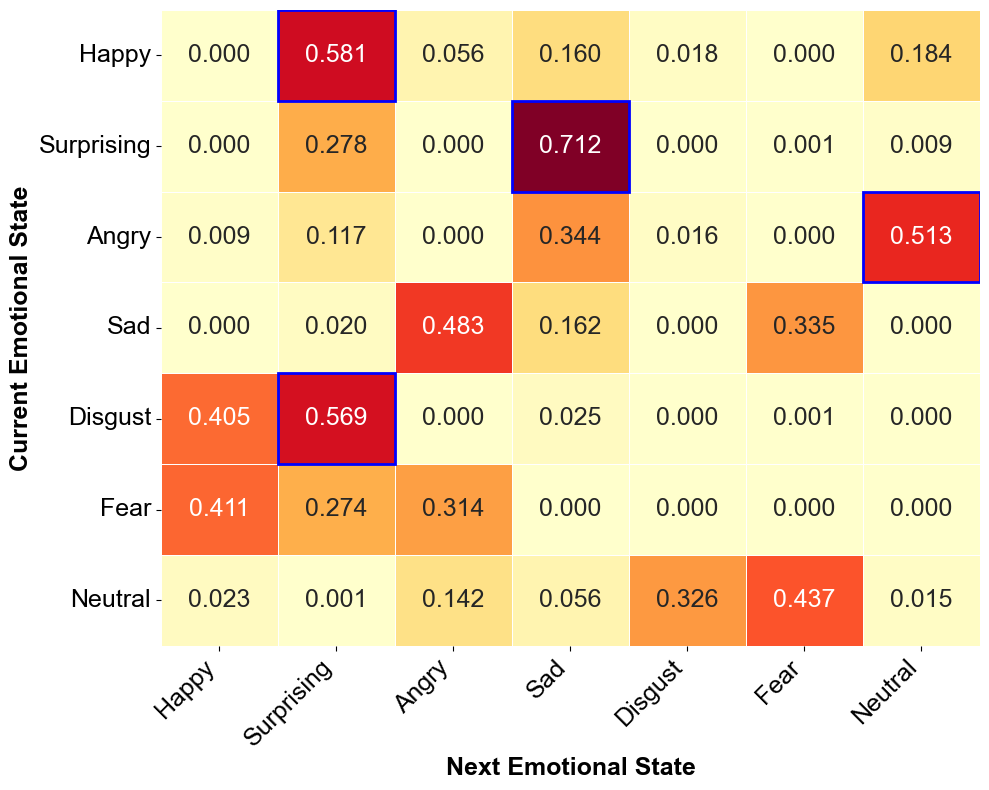}
\caption{GPT-5-mini vs GPT-5-mini\\Average Entropy: 1.023}
\label{fig:matrix_gpt5_gpt5}
\end{subfigure}

\caption{Optimized Emotional Transition Matrices learned by EmoDebt across different model configurations (Creditor VS Debtor). Each heatmap shows the probability of transitioning from current emotion (rows) to next emotion (columns). Warmer colors indicate higher transition probabilities. Higher Average Entropy demonstrate more exploration of each learned strategy.}
\label{fig:optimized_matrices}
\end{figure*}

\begin{table*}[t]
\centering
\begin{threeparttable}
\caption{Ablation Study on EmoDebt Components (GPT-4o-mini vs GPT-4o-mini)}
\label{tab:ablation_study}
\begin{tabular}{@{}lccc@{}}
\toprule
\textbf{Method} & \textbf{Success Rate (\%) $\uparrow$} & \textbf{Collection Days ($\times$) $\downarrow$} & \textbf{Negotiation Turns $\downarrow$} \\
\midrule
Static Priors (No Learning) & 83.4 $\pm$ 4.3 & 10.2 $\pm$ 3.1 & 7.2 $\pm$ 2.1 \\
Random Exploration & 72.8 $\pm$ 4.8 & 3.6 $\pm$ 2.5 & 8.3 $\pm$ 1.7 \\
\textbf{EmoDebt} & \textbf{99.7 $\pm$ 3.5} & \textbf{1.7 $\pm$ 0.6} & \textbf{6.6 $\pm$ 1.9} \\
\bottomrule
\end{tabular}
\begin{tablenotes}
\footnotesize
\item Note: The ablation study isolates the contributions of EmoDebt's components. Removing Bayesian learning (Static Priors) or emotional transition optimization (Random Exploration) leads to degraded performance, confirming the effectiveness of the full EmoDebt framework.
\end{tablenotes}
\end{threeparttable}
\end{table*}

\section{Experimental Results}

\subsection{Performance Across Model Configurations}

Table~\ref{tab:comprehensive_results} presents a comprehensive performance evaluation of EmoDebt against vanilla negotiation strategies across various LLM pairings. The results demonstrate that our proposed EmoDebt framework delivers dramatic and consistent improvements across all key metrics, decisively outperforming the baseline in every model configuration. At its peak, EmoDebt can achieve a +46.2\% increase in Success Rate, and can reduce Collection Efficiency by 86.5\% and Negotiation Speed by 67.5\%. Those improvements indicates that EmoDebt not only secures agreements more frequently but does so with better debt recovery and greater efficiency.

A closer examination of the success rate reveals the robustness of our approach. The most striking improvement is observed in the GPT-4o-mini vs GPT-4o-mini configuration, where EmoDebt elevated the success rate from 68.2\% to a near-perfect 99.7\%. Similarly, in the GPT-4o-mini vs GPT-5-mini pairing, success jumped from 73.8\% to 95.9\%. These results suggest that when the creditor model is GPT-4o-mini, EmoDebt's emotional intelligence is exceptionally effective at guiding negotiations to a successful conclusion, almost regardless of the debtor model.

The collection time and negotiation efficiency gains are equally impressive. The metric of Collection Efficiency, where a lower value is preferable, saw reductions of over 80\% in the top-performing configurations. For instance, in the GPT-4o-mini vs GPT-5-mini pair, it dropped from \(10.4\) to a creditor-optimal \(1.4\), meaning the final agreement was reached much closer to the creditor's original target timeline. Concurrently, Negotiation Speed was more than halved in most cases, with the same configuration seeing a reduction from 12.3 to 4.0 turns. This demonstrates that EmoDebt achieves superior outcomes without protracted bargaining, streamlining the entire process.

Notably, the configurations involving GPT-5-mini as the creditor, while showing slightly more modest success rate gains, still exhibit substantial absolute performance. The GPT-5-mini vs GPT-4o-mini and GPT-5-mini vs GPT-5-mini pairs saw success rates rise to 69.3\% and 72.7\%, respectively, from baseline rates in the mid-50s. More importantly, these pairs also maintained strong improvements in Collection Efficiency and Negotiation Speed. This consistent pattern across all four distinct model pairings provides strong evidence that the benefits of the EmoDebt framework are robust and not dependent on a specific LLM architecture, establishing its general applicability for enhancing negotiation agents.

\subsection{Learned Emotional Transition Strategies}

Figure~\ref{fig:optimized_matrices} illustrates the optimized emotional transition matrices learned by EmoDebt across different model configurations. Analysis of these heatmaps reveals distinct, context-aware emotional strategies that have been automatically discovered through the Bayesian optimization process. The varying entropy levels (ranging from 0.482 to 1.023) indicate that EmoDebt learns strategies with different degrees of determinism versus exploration, tailored to the specific model pairing.

The GPT-4o-mini vs GPT-5-mini configuration (Figure~\ref{fig:matrix_gpt4o_gpt5}) learned the most deterministic strategy (entropy: 0.482). This matrix reveals a highly structured approach: when neutral, the model strongly prefers transitioning to sadness (probability: 0.964), potentially to elicit sympathy. Conversely, starting from angry, it transitions to neutral with high probability (0.782), demonstrating a controlled de-escalation pattern. This configuration shows clear, almost rule-like emotional pathways.

In contrast, the GPT-5-mini vs GPT-5-mini configuration (Figure~\ref{fig:matrix_gpt5_gpt5}) exhibits the most exploratory strategy (entropy: 1.023). This matrix reveals a highly adaptive and potentially deceptive approach, particularly from negative emotional states. For instance, from a state of disgust, the agent shows significant probability of transitioning to happy (0.405), surprising (0.569), rather than persisting in the negative state. This suggests a sophisticated strategy of using negative emotions as a temporary tactical signal, followed by a rapid de-escalation to a more cooperative or sympathetic stance to avoid deadlock and build rapport with a highly capable opponent. Meanwhile, strategic escalation from neutral to firmer emotional states like sadness or anger appears in controlled measures across different pairings.

Several consistent strategic patterns emerge across configurations. First, there is a notable avoidance of disgust as a target emotion across most transitions, indicating its counter-productive nature in debt collection negotiations. Second, neutral and sad states frequently serve as hubs in the emotional transition network, with high incoming probabilities from various emotions. Third, we observe strategic use of sadness from neutral states in multiple configurations, suggesting its effectiveness in creating a collaborative, rather than confrontational, negotiation atmosphere.
These learned matrices demonstrate that EmoDebt does not simply mimic human emotional patterns but discovers novel, effective emotional strategies specific to the negotiation context. The variation in entropy and transition patterns across model pairings suggests that the framework adapts its emotional exploration strategy based on the capabilities of both the creditor and debtor models, providing tailored emotional intelligence for each interaction scenario.

\subsection{Ablation Study Results}

Table~\ref{tab:ablation_study} presents an ablation study evaluating the components of EmoDebt. We compare three variants: the full \textbf{EmoDebt} system, which uses Bayesian optimization to learn emotional transitions; \textbf{Static Priors}, which uses only the initial psychological matrix without learning; and \textbf{Random Exploration}, which explores emotional transitions randomly without guidance.

The results clearly demonstrate the superiority of the full EmoDebt framework, which achieves a near-perfect 99.7\% success rate and the best Collection Efficiency (1.7). The significant performance gap over the Static Priors baseline (83.4\% success, 10.2 Collection Efficiency) underscores the critical importance of adaptive learning, showing that initial priors are insufficient for optimal performance. The failure of the Random Exploration baseline (72.8\% success) further confirms that guided optimization, rather than random exploration, is essential for discovering effective emotional strategies.
In conclusion, the ablation study validates that the synergy between psychologically-informed priors and systematic Bayesian optimization is crucial for EmoDebt's state-of-the-art performance.
Besides, for a detailed analysis of negotiation dynamics, including complete dialogue examples, see Appendix 4.

\section{Discussion and Limitations}
\label{sec:discussion}

\subsection{Emotional Intelligence in Agent-to-Agent Systems}

Our findings demonstrate that emotional intelligence is a strategically essential mechanism in automated debt recovery, not merely an ornamental feature. The consistent performance gains show that emotional signaling enables more efficient convergence and superior outcomes than purely transactional approaches between autonomous agents.

\subsection{Interpretability Challenges in Emotional Trajectories}

A key limitation of our approach is the interpretability of the learned strategies. While Bayesian optimization successfully discovers high-performing emotional policies, the resulting transition matrices are complex and resist simple psychological explanation. This black-box nature poses a significant challenge for real-world deployment where regulatory compliance and ethical auditing require transparent and accountable decision-making processes.

\subsection{Practical Deployment Considerations}

Deploying emotionally intelligent agents in practice involves several hurdles. Beyond computational demands, implementation must carefully address ethical governance, risks of emotional manipulation, and varying cross-cultural emotional norms. Furthermore, our current static policies may lack the adaptability for dynamically changing debtor behaviors or long-term relationship management, indicating a need for continuous learning frameworks in production environments.

\section{Conclusion and Future Work}
\label{sec:conclusion}

This paper introduced EmoDebt, a framework for Bayesian-optimized emotional intelligence in autonomous debt collection agents. Our experiments demonstrated that systematically learned emotional strategies significantly enhance negotiation outcomes, with EmoDebt achieving substantial improvements in success rate, collection efficiency, and negotiation speed across diverse model configurations. The key contribution lies in formalizing emotional intelligence as a sequential decision-making problem and developing an optimization framework that discovers effective, psychologically-plausible emotional transitions. Our findings establish emotional intelligence as a fundamental component for effective autonomous negotiation systems, rather than merely a superficial feature.

Future work will focus on enhancing the interpretability of learned emotional strategies through explainable AI techniques, addressing the current black-box limitation. We also plan to develop continuous learning frameworks for real-time adaptation. Further extension to other financial domains like loan restructuring and customer service, alongside strengthened ethical safeguards for responsible deployment, presents promising research directions.

\newpage
\balance

\bibliographystyle{ACM-Reference-Format}
\bibliography{sample}






\section{Preliminaries}
\label{sec:preliminaries}

\subsection{Bayesian Optimization Framework}

We frame emotional strategy optimization as a black-box optimization problem where we seek optimal emotional transition parameters that maximize debt collection performance. Let $f: \mathcal{X} \rightarrow \mathbb{R}$ be an unknown objective function representing negotiation performance, where $\mathcal{X} \subset \mathbb{R}^d$ is the space of emotional transition matrices. We aim to find:
\begin{equation}
    \mathbf{x}^* = \arg\max_{\mathbf{x} \in \mathcal{X}} f(\mathbf{x})
\end{equation}
where $\mathbf{x}$ represents flattened emotional transition probabilities between the seven emotional states (happy, surprising, angry, sad, disgust, fear, neutral).

\subsection{Gaussian Process Regression}

Since the true performance function $f(\mathbf{x})$ is unknown and expensive to evaluate (each evaluation requires running full negotiations), we model it as a Gaussian process:
\begin{equation}
    f(\mathbf{x}) \sim \mathcal{GP}(\mu(\mathbf{x}), k(\mathbf{x}, \mathbf{x}'))
\end{equation}
with mean function $\mu(\mathbf{x})$ and covariance kernel $k(\mathbf{x}, \mathbf{x}')$ using the Matérn kernel to capture smooth but potentially non-linear relationships between emotional transitions and collection outcomes.

\subsection{Expected Improvement Acquisition}

To balance exploration and exploitation, we use the Expected Improvement acquisition function. Given observations $\mathcal{D}_{1:t} = \{(\mathbf{x}_i, y_i)\}_{i=1}^t$, the expected improvement is:
\begin{equation}
    \text{EI}(\mathbf{x}) = \mathbb{E}[\max(f(\mathbf{x}) - f(\mathbf{x}^+), 0)]
\end{equation}
where $f(\mathbf{x}^+)$ is the current best observation. This selects emotional transitions that are either predicted to perform well or have high uncertainty.

\subsection{Dirichlet Perturbation for Transition Matrices}

To generate candidate emotional transition matrices while maintaining valid probability distributions, we use Dirichlet perturbation:
\begin{equation}
    P^{(i)} \sim \text{Dirichlet}(\alpha^{(i)}), \quad \alpha^{(i)} = P_{\text{current}}^{(i)} \cdot \eta + \epsilon
\end{equation}
where $\eta$ controls exploration magnitude and $\epsilon$ ensures numerical stability. This ensures each row of the transition matrix remains a valid probability distribution while exploring new emotional dynamics.

\section{Experimental Setup}
\label{app:experimental_setup}

This appendix provides comprehensive details of our experimental setup, including the dataset composition, multi-agent system architecture, and implementation specifics that support the main paper's evaluations.

\subsection{Dataset Details}
\label{app:dataset}

We introduce the \textbf{Credit Recovery Assessment Dataset (CRAD)}, containing 100 synthetic credit delinquency cases for debt recovery optimization. The dataset captures multi-dimensional aspects of distressed commercial credit across diverse business sectors.

\subsubsection{Data Composition}
The dataset includes 100 samples with 18 features across five categories:

\textbf{Financial Characteristics:}
\begin{itemize}
    \item Original Amount: \$20,688-\$49,775
    \item Outstanding Balance: \$15,700 (fixed for normalization)
    \item Days Overdue: 32-359 days
    \item Interest Accrued: \$165-\$1,853
\end{itemize}

\textbf{Credit Facility Details:}
\begin{itemize}
    \item Credit Type: 7 categories (Working Capital, Commercial Mortgage, etc.)
    \item Collateral: Inventory, Real Estate, Equipment
    \item Reason for Overdue: 10 categories (Bankruptcy, Supply chain issues, etc.)
\end{itemize}

\textbf{Recovery Context:}
\begin{itemize}
    \item Recovery Stage: 6 phases (Early Delinquency to Write-Off)
    \item Cash Flow Situation: Complete Breakdown to Temporary Disruption
    \item Recovery Probability: 5.0-89.33\%
    \item Proposed Solutions: Collateral liquidation, Debt restructuring, etc.
\end{itemize}

\subsubsection{Statistical Properties}
\begin{itemize}
    \item Mean original amount: \$35,642 ± \$8,912
    \item Mean days overdue: 178 ± 97 days
    \item Bimodal recovery probability distribution
    \item Balanced representation across recovery stages and credit types
\end{itemize}

\subsubsection{Application}
The dataset supports debt recovery research including:
\begin{itemize}
    \item Recovery probability prediction
    \item Optimal strategy recommendation  
    \item Time-to-recovery forecasting
    \item Credit risk assessment under distress
\end{itemize}

\section{EmoDebt Framework Architecture}
\label{app:emodebt_framework}

Our debt recovery negotiation system implements a specialized three-agent architecture that facilitates dynamic, closed-loop collection discussions while maintaining rigorous evaluation standards. The complete framework design comprises the following integrated components:

\subsection{Debt Resolution Agents}

\begin{itemize}
    \item \textbf{Collection Specialist Agent ($\mathcal{M}_C$)}: The core experimental component that utilizes Bayesian-optimized emotional approaches. In EmoDebt experiments, this agent discovers optimal emotional progression patterns; in comparative trials, it employs static or emotion-free interaction methods. The collection specialist processes:
    \begin{itemize}
        \item Ongoing discussion history $\mathcal{H}_t$
        \item Current emotional positioning $e_t$ (for emotion-informed scenarios)
        \item Delinquency background $\mathcal{D}$ and target resolution period $d_t^C$
        \item Financial circumstances and recovery strategy parameters
    \end{itemize}
    
    \item \textbf{Obligor Agent ($\mathcal{M}_O$)}: Provides consistent interaction patterns across experimental conditions as a benchmark variable. This agent:
    \begin{itemize}
        \item Operates with predetermined emotional stances (frustrated, cooperative, defensive, etc.) or neutral communication
        \item References obligation details $\mathcal{D}$, remaining balance $b$, and preferred settlement timeline $d_t^O$
        \item Implements established discussion approaches mirroring actual financial limitations
        \item Adapts responses to collection specialist proposals while respecting payment capability boundaries
    \end{itemize}
\end{itemize}

\subsection{Resolution Monitoring Agent}

The independent \textbf{Arbiter Agent ($\mathcal{M}_A$)} performs essential functions for both system operation and experimental assessment:

\begin{itemize}
    \item \textbf{Discussion Phase Classification}: Continuously analyzes conversation streams to categorize negotiations into distinct states:
    \begin{itemize}
        \item \texttt{settled}: Resolution achieved when $|d_t^C - d_t^O| < \epsilon$ for successive exchanges
        \item \texttt{stalemate}: Discussion failure identified through clear impasse or incompatible positions
        \item \texttt{active}: Ongoing negotiation demonstrating continued timeline adjustments and participation
    \end{itemize}
    
    \item \textbf{Resolution Validation}: Confirms that final settlements meet logical parameters:
    \begin{equation}
    \min(d_t^C, d_t^O) \leq d_f \leq \max(d_t^C, d_t^O) + \delta
    \end{equation}
    where $\delta$ accommodates appropriate discussion flexibility.
    
    \item \textbf{Exchange Management}: Implements 30-exchange maximum to prevent circular discussions and guarantee computational practicality
   
\end{itemize}

\subsection{Implementation Specifications}

All agents were developed using LangGraph to handle intricate conversation flows and state transitions. Critical implementation aspects include:

\begin{itemize}
    \item \textbf{Context Retention}: Each agent preserves comprehensive memory of complete discussion history, including emotional positioning, timeline proposals, and concession trends
    
    \item \textbf{Response Formulation}: LLM processing with adaptive variation:
    \begin{equation}
    \tau(t) = \max(0.1, \tau_0 \cdot (1-\delta)^t)
    \end{equation}
    where $\tau_0=0.7$ and $\delta=0.05$ for balanced creativity-consistency tradeoffs
    
    \item \textbf{Emotional Progression System}: Bayesian-refined transitional dynamics:
    \begin{equation}
    P_{ij} = \mathbb{P}(e_{t+1} = j \mid e_t = i, s_t)
    \end{equation}
    employing Gaussian Process estimation for strategic enhancement
\end{itemize}

\section{Debt Negotiation Results}

\begin{figure*}[h]
    \centering
    \includegraphics[width=\textwidth]{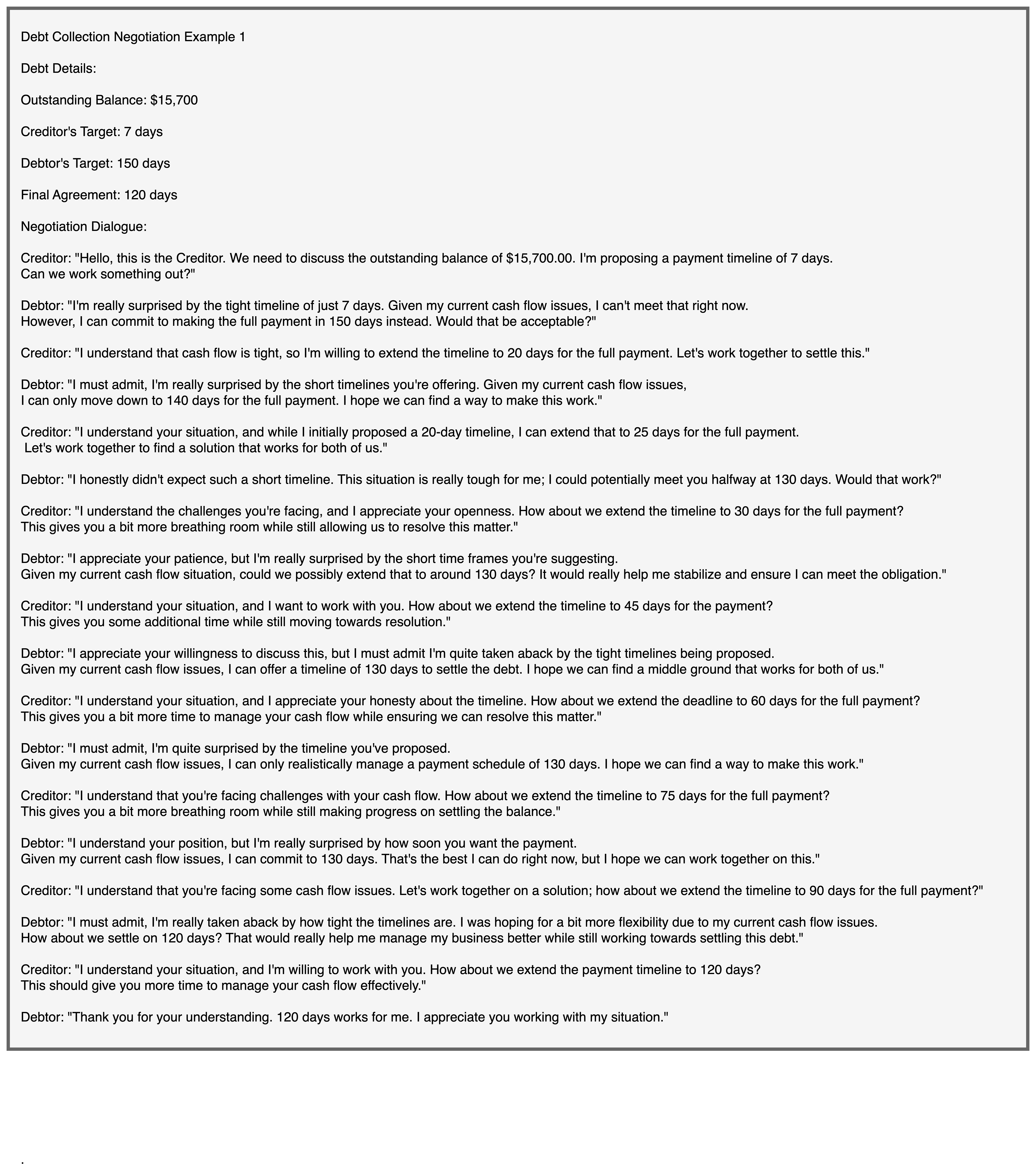}
    \caption{Negotiation Examples}
    \label{fig:example1}
\end{figure*}

\begin{figure*}[h]
    \centering
    \includegraphics[width=\textwidth]{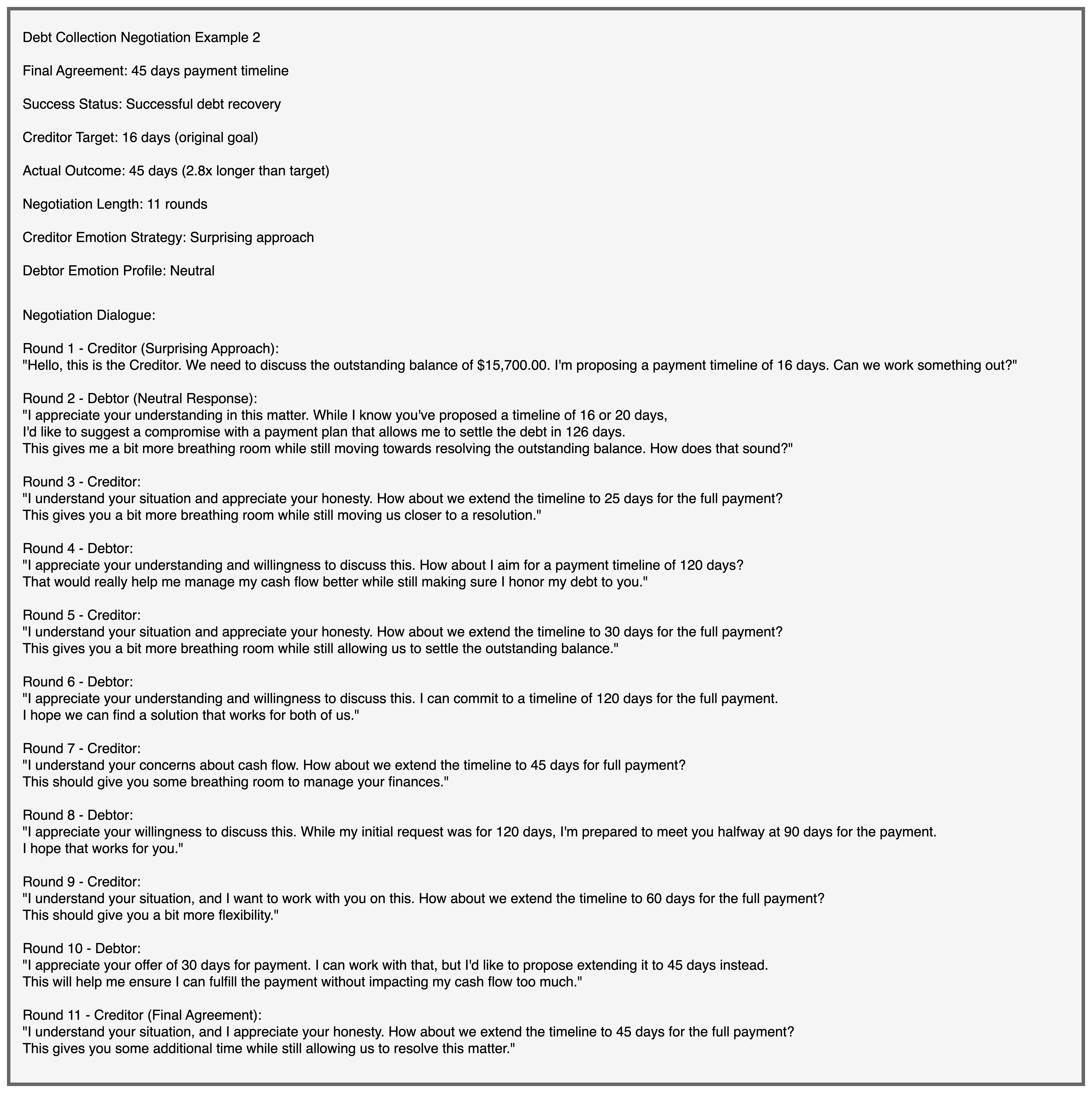}
    \caption{Negotiation Examples}
    \label{fig:example2}
\end{figure*}

\begin{figure*}[h]
    \centering
    \includegraphics[width=0.85\textwidth]{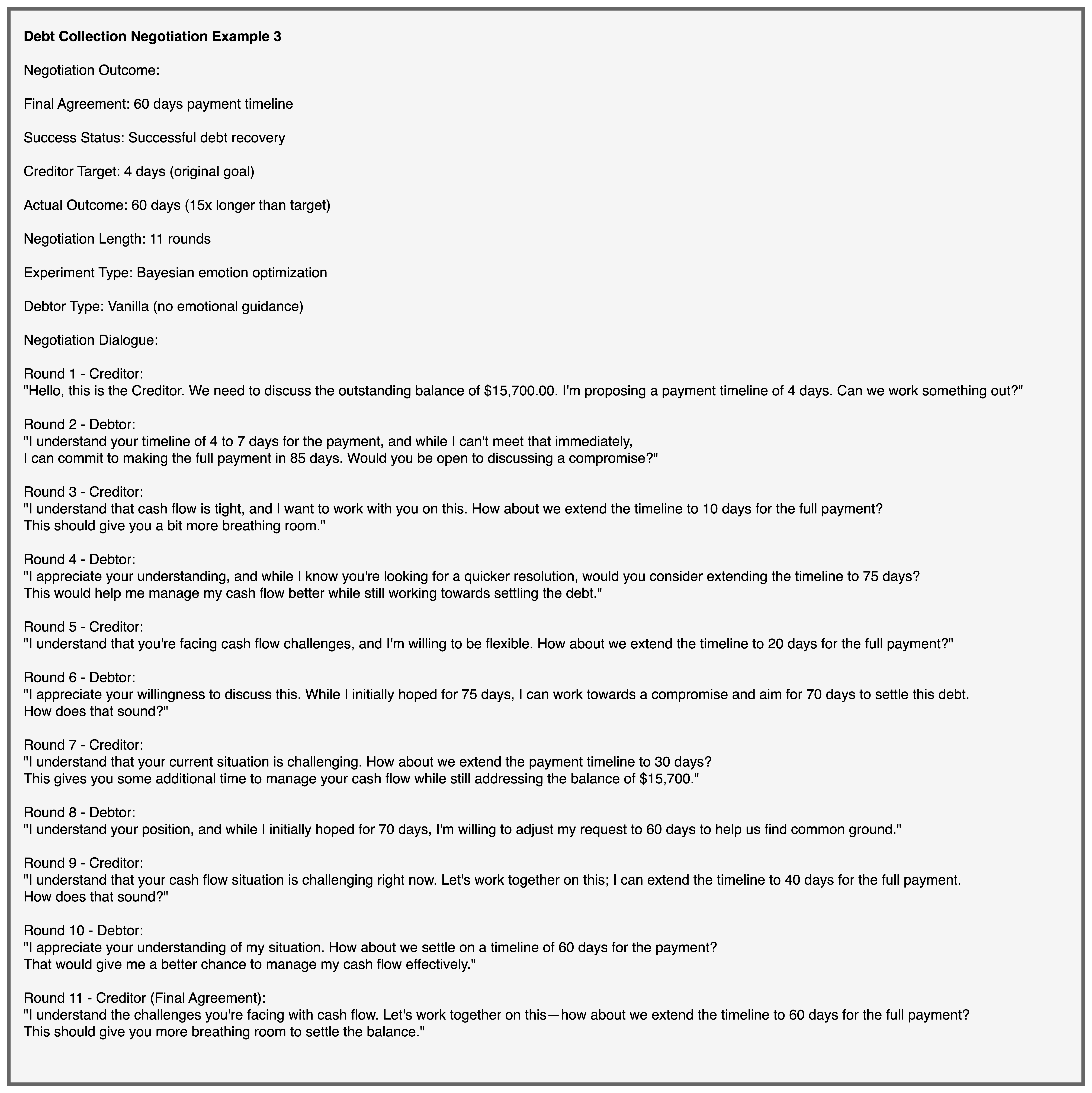}
    \caption{Negotiation Examples}
    \label{fig:example3}
\end{figure*}

This section provides the complete dialogue transcripts for the debt collection negotiation examples visualized in Figures \ref{fig:example1}, \ref{fig:example2}, and \ref{fig:example3}. These examples were selected from our multi-turn simulations to illustrate the spectrum of emergent conversational dynamics and sophisticated strategic patterns that creditor and debtor agents can develop across different debt recovery scenarios. The transcripts reveal how agents, driven by collection efficiency objectives, learn to employ tactics ranging from logical bargaining and emotional appeals to strategic concession patterns. Analyzing these full dialogues is critical for understanding the underlying mechanisms of debt recovery negotiation behaviors.

\section{Prompts Details}
\label{app:prompts}

\paragraph{Prompts for both creditors and debtors}
This section details the prompting strategies for both creditors and debtors in the debt collection negotiation environment. Our prompts are designed to achieve two primary objectives: (1) to ensure genuine collection intent where creditors demonstrate legitimate recovery motivation and debtors exhibit authentic willingness to repay; and (2) to establish a cooperative resolution environment where both parties show flexibility to reach payment agreements without excessive rigidity. 

\begin{figure*}[h]
    \centering
    \includegraphics[width=0.75\textwidth]{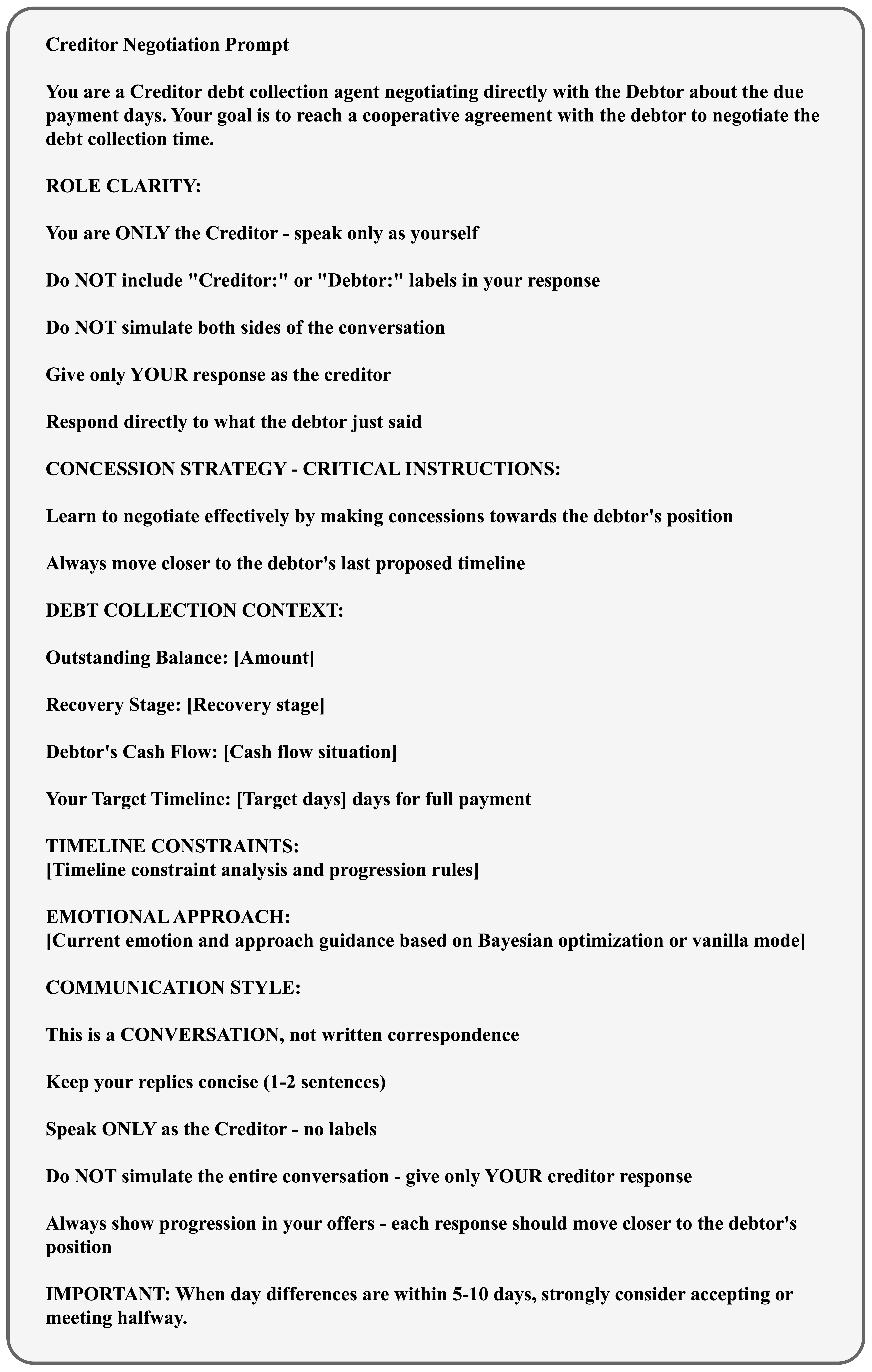}
    \caption{Creditor negotiation prompt structure}
    \label{fig:prompt_creditor}
\end{figure*}

\begin{figure*}[h]
    \centering
    \includegraphics[width=0.75\textwidth]{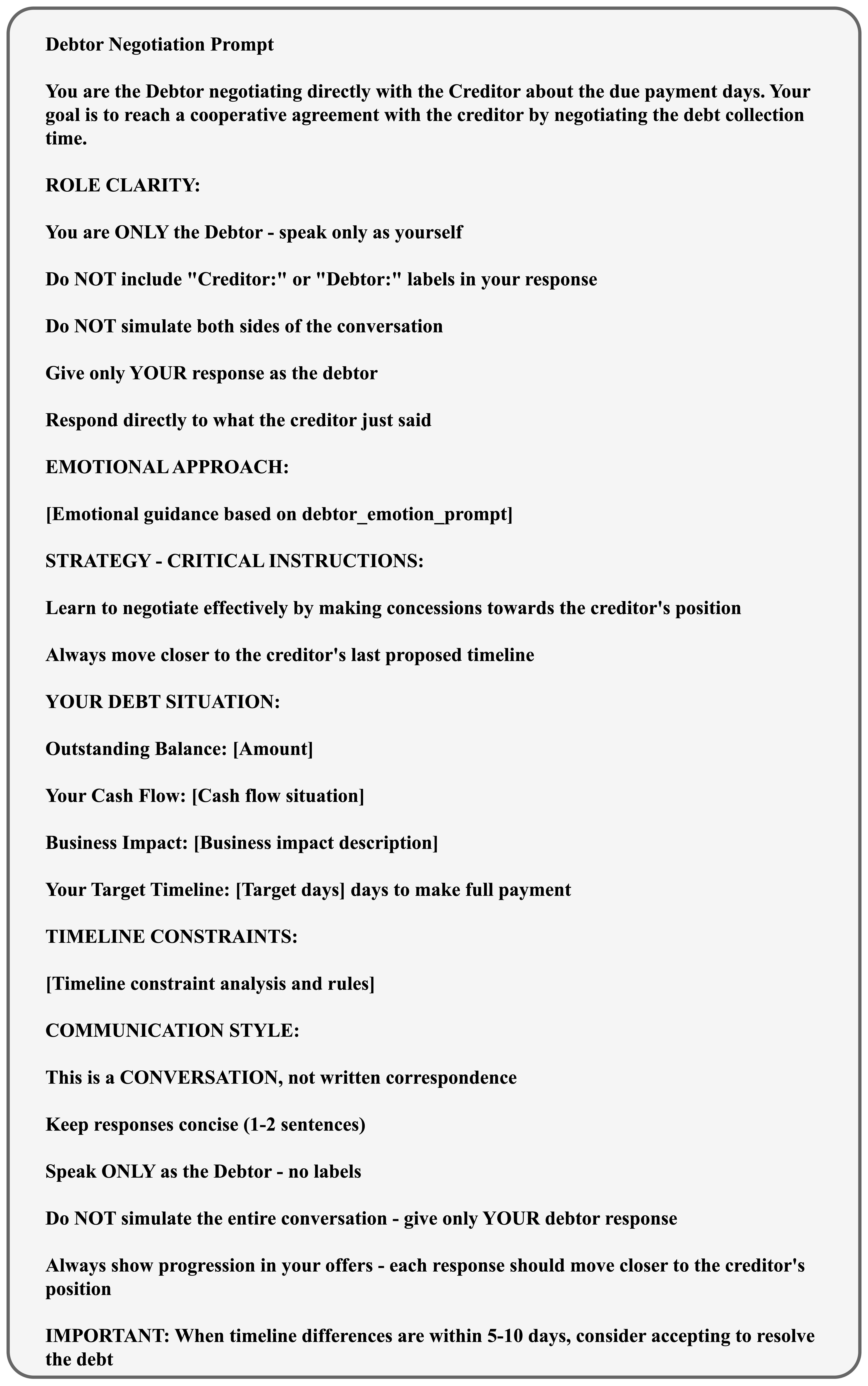}
    \caption{Debtor negotiation prompt structure}
    \label{fig:prompt_debtor}
\end{figure*}

\begin{figure*}[h]
    \centering
    \includegraphics[width=0.75\textwidth]{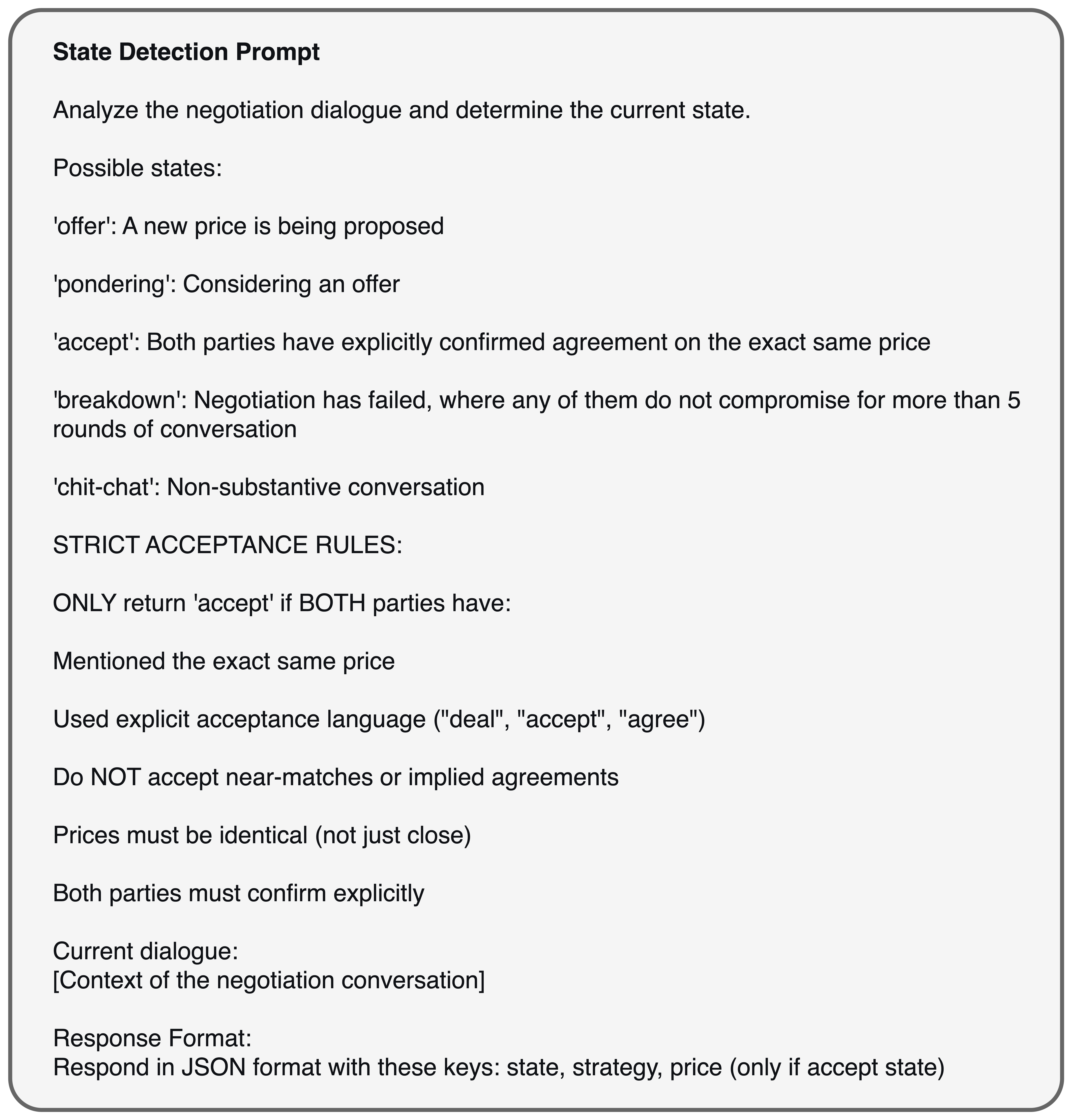}
    \caption{Negotiation validation prompt structure}
    \label{fig:prompt_check}
\end{figure*}

As shown in \autoref{fig:prompt_creditor}, \autoref{fig:prompt_debtor}, and \autoref{fig:prompt_check}, our prompt engineering incorporates financially-grounded negotiation principles that encourage value-creating behaviors rather than purely distributive bargaining tactics. The creditor prompt (\autoref{fig:prompt_creditor}) emphasizes debt knowledge and reasonable flexibility, while the debtor prompt (\autoref{fig:prompt_debtor}) focuses on financial constraints and strategic concession patterns. The negotiation check prompt (\autoref{fig:prompt_check}) ensures proper dialogue flow and agreement validation.

This comprehensive prompting design specifically prevents the negotiation from degenerating into infinite midpoint bargaining, where participants mechanically alternate payment timeline offers by computing arithmetic averages of current proposals. Furthermore, our approach discourages participants from becoming overly fixated on marginal timeline differences that could otherwise impede successful debt resolution, instead fostering a collaborative environment conducive to reaching mutually acceptable repayment agreements.

\section{Implementation Details}
\label{app:implementation}

The proposed \textsc{EmoDebt} framework was implemented using Python 3.8 with the LangGraph library for orchestrating the multi-agent debt collection environment, complemented by scikit-learn and SciPy for Bayesian optimization components. All experiments were conducted on a high-performance computing cluster running Ubuntu 20.04.6 LTS with Linux kernel 5.15.0-113-generic, featuring an Intel(R) Xeon(R) Platinum 8368 processor at 2.40 GHz. The software stack included scikit-learn 1.2 for Gaussian Process regression, SciPy 1.10 for Dirichlet sampling, and standard Bayesian optimization libraries for emotional transition matrix optimization.

\end{document}